\documentclass[conference]{IEEEtran}
\IEEEoverridecommandlockouts

\usepackage{cite}
\usepackage{amsmath,amssymb,amsfonts}
\usepackage{algorithmic}
\usepackage{graphicx}
\usepackage{textcomp}
\usepackage{xcolor}
\usepackage{siunitx} 
\usepackage{booktabs} 
\usepackage{adjustbox} 
\usepackage{multirow}

\usepackage{comment}

\def\BibTeX{{\rm B\kern-.05em{\sc i\kern-.025em b}\kern-.08em
    T\kern-.1667em\lower.7ex\hbox{E}\kern-.125emX}}
\begin{document}

\title{Interactive Masked Image Modeling for Multimodal Object Detection in Remote Sensing
}


\author{
    Minh-Duc VU, Zuheng MING, Fangchen FENG, Bissmella BAHADURI, Anissa MOKRAOUI\\
    L2TI Laboratory, University Sorbonne Paris Nord, Villetaneuse, France\\
    minh-duc.vu@epita.fr, \{zuheng.ming, fangchen.feng, bissmella.bahaduri, anissa.mokraoui\}@univ-paris13.fr
}

\maketitle

\begin{abstract}

Object detection in remote sensing imagery plays a vital role in various Earth observation applications. However, unlike object detection in natural scene images, this task is particularly challenging due to the abundance of small, often barely visible objects across diverse terrains. To address these challenges, multimodal learning can be used to integrate features from different data modalities, thereby improving detection accuracy. Nonetheless, the performance of multimodal learning is often constrained by the limited size of labeled datasets. In this paper, we propose to use Masked Image Modeling (MIM) as a pre-training technique, leveraging self-supervised learning on unlabeled data to enhance detection performance. However, conventional MIM such as MAE which uses masked tokens without any contextual information, struggles to capture the fine-grained details due to a lack of interactions with other parts of image. To address this, we propose a new interactive MIM method that can establish interactions between different tokens, which is particularly beneficial for object detection in remote sensing. The extensive ablation studies and evluation  demonstrate the effectiveness of our approach.

\end{abstract}

\begin{IEEEkeywords}
Self-supervised learning, multimodal learning, object detection, remote sensing image.
\end{IEEEkeywords}

\section{Introduction}
Object detection in Remote Sensing Images (RSI) including aerial images is a critical task enabling the identification and localization of objects within satellite or aerial imagery. It has numerous applications for Earth Observation (EO) such as environmental monitoring, climate change, urban planning, and military surveillance~\cite{gui2024remote}. Object detection in aerial imagery is a complex task, made even more challenging by several critical factors. These include the limited availability of annotated data compared to standard imagery~\cite{le2022improving}, the smaller size of objects in aerial environments~\cite{xu2022detecting}, and the distinct top-down perspective inherent to aerial observations.

To overcome these challenges, complementary information from alternative modalities, such as infrared (IR) which allows us to see through certain obstructions like smoke or fog, can be leveraged~\cite{sharma2020yolors}. Both IR and RGB sensors are widely used in aerial vehicles and Earth observation satellites. By combining these spectral channels, multispectral images are generated, providing far richer information than RGB images alone~\cite{li2022deep}. This enhanced data stream has the potential to significantly improve the precision and accuracy of object detection in aerial imagery. However, detection performance in multimodal scenarios is often limited by the small size of available datasets, as collecting multimodal data is inherently challenging. Additionally, multimodal architectures are generally more complex than single-modal ones and require larger datasets for effective training.


\begin{figure}
    \centering
    \includegraphics[width=\linewidth]{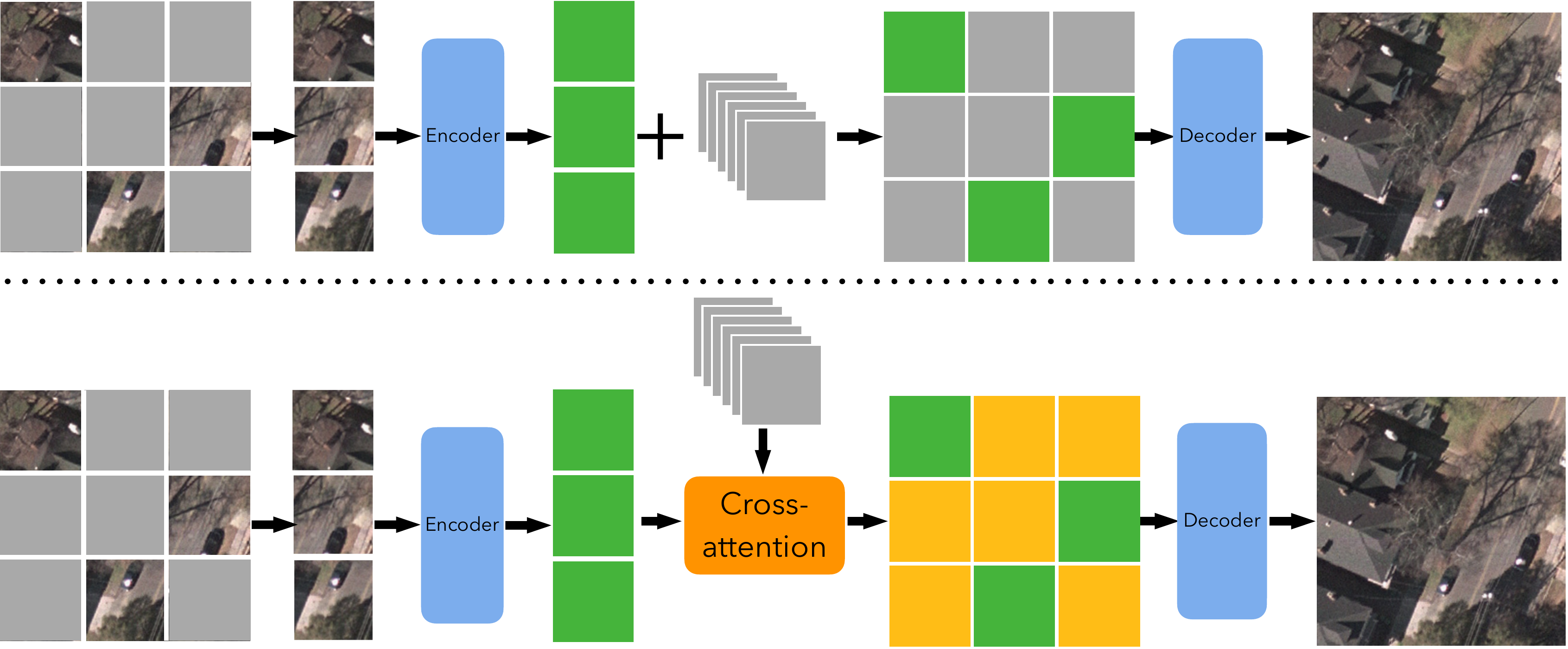}
    \caption{Interactive masked image modeling for self-supervised pre-training. The top is the conventional masked image modeling such as MAE\cite{he2022masked}. The bottom is the interactive masked image modeling, in which a cross-attention is introduced to create the interaction between unmasked tokens and masked tokens. The features of unmasked token from encoder (green squares) are merged with the features of masked token from the cross-attention module (orange squares) to reconstruct the masked images.}
    \label{fig:CAD}
\end{figure}

To mitigate this issue, self-supervised learning (SSL) has gained prominence as a powerful deep learning approach, particularly for tasks where labeled data is scarce or expensive to obtain. SSL leverages unlabeled data by creating pretext tasks that enables the model to learn generic and comprehensive features of the images~\cite{geiping2023cookbook}. A model pretrained on a pretext task can be more efficiently adapted to downstream tasks through fine-tuning with limited domain-specific labeled data. As one of the most widely used SSL method, Masked Image Modeling~\cite{xie2022simmim} (MIM) allows the model to understand the underlying concepts of an image by predicting the masked portions of an input image (see Figure \ref{fig:CAD}), thereby forcing the model to develop a deeper understanding of the image structure. This approach is particularly effective in scenarios where images contain significant amounts of contextual information, such as aerial imagery or remote sensing~\cite{sun2022ringmo}. 

Nevertheless, conventional MIM methods such as MAE \cite{he2022masked} reconstruct masked images using the features of both unmasked and masked tokens. The masked tokens are typically set to null, containing no information. Consequently, the interaction between different parts of the image, such as between unmasked and masked tokens, is often interrupted during reconstruction. While this approach can help the model learn the global context, it may not always capture fine-grained details that are crucial for detecting small or densely packed objects in remote sensing images, as the surrounding contextual information of an unmasked token cannot be provided by neighboring masked tokens. Due to the lack of interaction between unmasked and masked tokens, standard MIM methods may also struggle with variations in object size and contextual complexity, such as diverse terrains in remote sensing images, as they may not fully capture these variations in their representations.  In this work, we propose an interactive MIM, in which a cross-attention module is introduced to the standard MIM to create dependencies between unmasked and masked tokens. The generated features derived from the masked tokens via the cross-attention module are then merged with the features of the unmasked tokens to reconstruct the masked images. This approach encourages the encoder to learn a more holistic understanding of the image content, which benefits downstream object detection tasks in remote sensing images.

To summarize, the main contributions of this work are:
\begin{enumerate} 
    \item We propose using multimodal SSL for object detection in remote sensing images to address data scarcity by leveraging the rich unlabeled data from different sources.
    \item To address the limitations of traditional MIM, we propose a new interactive masked image modeling method that better supports downstream object detection tasks in remote sensing images.
    \item The extensive experiments including ablation studies and overall evaluations demonstrate the effectiveness of the proposed approach in both single-modality and multimodal settings.
\end{enumerate}

\section{Related Works}

Since its introduction, Masked Image Modeling (MIM) has gained significant popularity as a pre-training strategy, leading to extensive research aimed at improving its effectiveness. Several studies have focused on enhancing the masking process. In distillation-based MIM, for example, the authors of~\cite{kakogeorgiou2022hide} introduced a teacher transformer encoder that generates an attention map, guiding the masking process for the student model. Similarly, the authors of~\cite{liu2023good} used a self-attention mechanism to extract semantic information during training, which informs the masking strategy. A comparable approach is found in~\cite{objectCentric}, specifically for remote sensing images, where the authors developed an object-centric data generator to automatically configure pre-training data based on objects within the imagery. By masking regions centered around objects, this method encourages models to learn richer, object-specific features, outperforming traditional MIM techniques.

In addition to masking strategies, some works focus on the relationship between masked and unmasked tokens. The authors of~\cite{chen2024context} argue that it is beneficial to constrain predicted representations to the encoded representation space. Meanwhile, the authors of~\cite{xiong2024efficientsam} proposed a cross-attention decoder to combine masked and unmasked tokens—this work serves as the inspiration for our paper. Building on this approach, we extend its application to the more complex multimodal setting. One key improvement we introduce, as demonstrated through experiments, is that the query in the cross-attention decoder can be either masked or unmasked tokens, adding flexibility to the model. Further details are provided in the next section. Other relevant research includes foundation models for remote sensing images, also based on MIM~\cite{sun2022ringmo}.



\section{Methods}
\subsection{Overall architecture}
\begin{figure}[ht]
    \centering
    \includegraphics[width=\linewidth]{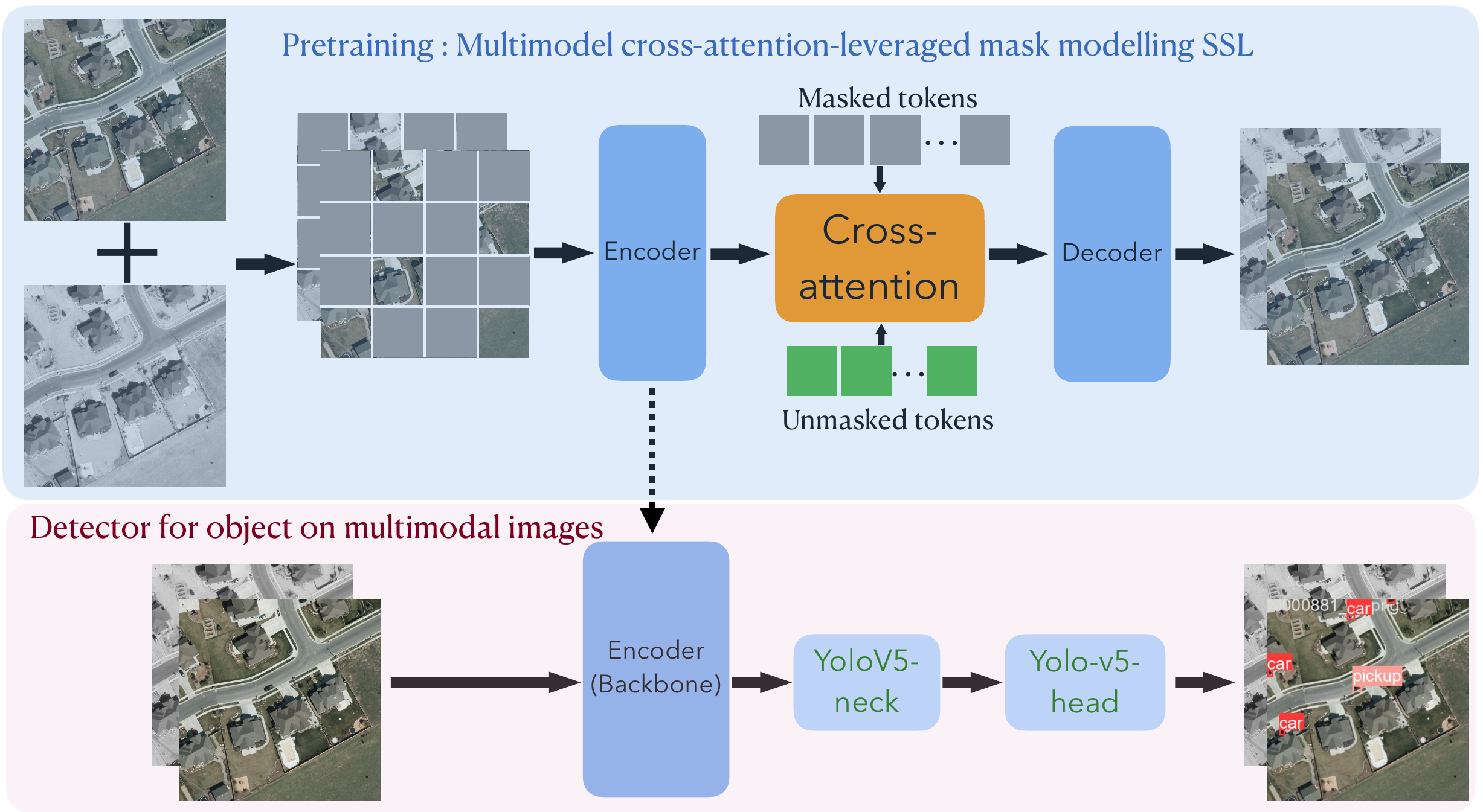}
    \caption{Overview of our framework. Our proposed framework consists of two stages: pre-training (top) on the AVIID or DIOR datasets, and fine-tuning (bottom) on VEDAI. During pre-training, the output features of unmasked tokens from the encoder, merged with the output features of masked tokens from the cross-attention module, are used to reconstruct the multimodal images. After pre-training, the decoder is discarded, and the pre-trained encoder serves as the image encoder for fine-tuning on VEDAI.}
    \label{fig:full-process}
\end{figure}
Figure \ref{fig:full-process} shows the overall architecture of our proposed multimodal cross-attention Masked Image Modeling (MIM) for object detection in remote sensing images. Similar to conventional SSL-based modeling \cite{chen2020simple}, our framework consists of two stages: pre-training and downstream task fine-tuning. The pretrained encoder in the first stage is adapted to downstream tasks, such as object detection, through fine-tuning. In the pre-training stage, the images of different modalities are first concatenated and masked. The unmasked tokens are fed to the encoder to obtain their features.  Unlike the MAE \cite{he2022masked}, one of the most widely used MIM methods for pre-training, we introduce a cross-attention module that uses the features of unmasked tokens as anchors to infer meaningful features of masked tokens that contains contextual information. The features of unmasked tokens and those derived from masked tokens are then input to the decoder to reconstruct the masked multimodal images, such as RGB-IR images. The details of the proposed cross-attention MIM will be elaborated in the following section. Here, we use the Swin Transformer \cite{liu2021swin} as the encoder, and YOLOv5 \cite{cao2023swin} as the detection neck/head in the downstream object detection task.

\subsection{Cross Attention MIM}
As shown in Figure \ref{fig:CAD}, we designed a cross-attention module to generate features from the masked tokens by creating dependencies between the masked and unmasked areas. Unlike conventional MIM using masked tokens (represented by the gray squares in Figure \ref{fig:CAD}), the proposed cross-attention MIM merges the output features of unmasked tokens (the green squares in Figure \ref{fig:CAD}) from the encoder with the features of masked tokens from the cross-attention module (the orange squares in Figure \ref{fig:CAD}) to reconstruct the image. The features from the masked tokens can capture the surrounding context of the unmasked tokens, which is lost in the original masked tokens set as null. This approach allows the encoder to be trained more effectively, capturing more generic and comprehensive features of the image, which benefits downstream tasks.


The cross-attention mechanism can be formulated as follows:
\begin{equation}
    \text{Attention}(Q_m, K_u, V_u) = \text{softmax} \left( \frac{Q_m K_u^T}{\sqrt{d_k}} \right) V_u
\end{equation}

Where:
\begin{itemize}
  \item $Q_m$: Query matrix for masked tokens
  \item $K_u$: Key matrix for unmasked tokens
  \item $V_u$: Value matrix for unmasked tokens
  \item $d_k$: Dimension of the key vectors
\end{itemize}
The obtained attention scores are used as the features of masked tokens to reconstruct the masked images.

\subsection{Reconstruction loss}
We employ the {$l_2$-norm} distance between the pixels of the original image and the reconstructed image to as the reconstruction loss, as described by the following equation:

\begin{equation}
    \begin{split}
    L_{W_e, W_d} = \frac{1}{N} \cdot \sum_{j = 1}^N ||\mathbf{x} - f^c(\mathbf{x})||^2,
    \end{split}
\end{equation}

where $W_e$ and $W_d$ represent the weights of the encoder and decoder respectively. Here, $N$ denotes the total number of samples, $\mathbf{x}$ is the original image, and $f^{c}(\mathbf{x})$ is the reconstructed image generated by the model $f^{c}(\mathbf{\cdot})$. When inputting multimodal images such as RGB-IR image pairs, the IR image is concatenated with RGB image in a channel-wise manner to form the new input for the model, which can be given by:
\begin{equation}
    \mathbf{x} = \mathbf{x}_{rgb} \oplus \mathbf{x}_{ir},
\end{equation}
where $\mathbf{x}_{rgb}$ is RGB image, $\mathbf{x}_{ir}$ is IR image and $\mathbf{x}$ is the multimodal input for the model, $\oplus$ denotes the concatenation of images in a channel-wise manner.


\section{Experiments}
\subsection{Experiments settings}
We apply our proposed self-supervised methods to three remote sensing image datasets for pre-training: VEDAI \cite{razakarivony2016vehicle}, DIOR \cite{li2020object}, and AVIID \cite{han2023aerial}. VEDAI consists of 1,246 images in both 1024×1024 and 512×512 resolutions. Each image has 4 channels: RGB and IR. DIOR is a large-scale remote sensing dataset containing over 23,000 high-resolution RGB images with approximately 200,000 annotated object instances across 20 different classes. AVIID is composed of 3 parts, but only part 3 consists of aerial images.AVIID-3 contains 1,280 pairs of RGB-IR images at a 512 × 512 resolution and has been used in this work for pre-training. For the downstream task, we only use the VEDAI dataset for training and validation.The dataset is divided into 10 folds and we use the first folder for the ablation studies, and all 10 folders for the overall evaluation.  The mAP (mean Average Precision) at 0.5 IoU, i.e. MAP@.5, is used as accuracy metrics to evaluate the object detection performance. We have used an AdamW optimizer with a base learning rate of 1e-5 and 0.005 weight decay. 8 Nvidia Tesla V100 GPUs were used for the training.

\subsection{Experiment results}


\begin{table*}[ht]
    \caption{Class-wise overall evaluation of the proposed methods for object detection in remote sensing images.}
    \centering
    \scriptsize
    \small
    \renewcommand{\arraystretch}{0.5} %
    \begin{adjustbox}{max width=\textwidth}
    \begin{tabular}{@{}c|c * {9}{|c}S[table-format=3.0]@{}}
        \toprule
        
        Model & Modality & {Car} & {Pickup} & {Camping} & {Truck} & {Other} & {Tractor} & {Boat} & {Van} & {mAP.5}\\
        \midrule
        \multirow{2}{*}{YOLOv3}\cite{redmon2018yolov3} & RGB & 0.83 & 0.71 & 0.69 & 0.59 & 0.48 & 0.67 & 0.33 & 0.55 & 0.61\\
        & RGB + IR & 0.84 & 0.72 & 0.67 & 0.62 & 0.43 & 0.65 & 0.37 & 0.58 & 0.61\\
        \midrule
        \multirow{2}{*}{YOLOv4}\cite{bochkovskiy2020yolov4} & RGB & 0.84 & 0.73 & 0.71 & 0.59 & {0.52} & 0.66 & 0.34 & 0.60 & 0.62\\
        & RGB + IR & 0.85 & 0.73 & \textbf{0.72} & 0.63 & 0.49 & 0.69 & 0.34 & 0.55 & 0.62\\
        \midrule
        
        \multirow{2}{*}{YOLOv5s}\cite{ultralytics_yolov5} & RGB & 0.80 & 0.68 & 0.66 & 0.51 & 0.46 & 0.64 & 0.22 & 0.41 & 0.55\\
        & RGB + IR & 0.81 & 0.68 & 0.69 & 0.55 & 0.47 & 0.64 & 0.24 & 0.46 & 0.57\\
        \midrule
        
        \multirow{2}{*}{YOLOv5m}\cite{ultralytics_yolov5} & RGB & 0.81 & 0.70 & 0.66 & 0.54 & 0.47 & 0.67 & 0.36 & 0.50 & 0.59\\
        & RGB + IR & 0.83 & 0.72 & 0.68 & 0.59 & 0.46 & 0.66 & 0.33 & 0.57 & 0.61\\
        \midrule
        
        \multirow{2}{*}{YOLOv5l}\cite{ultralytics_yolov5} & RGB & 0.81 & 0.72 & 0.68 & 0.57 & 0.46 & 0.71 & 0.36 & 0.55 & 0.61\\
        & RGB + IR & 0.83 & 0.72 & 0.70 & 0.64 & 0.48 & 0.63 & 0.40 & 0.56 & 0.62\\
        \midrule
        
        \multirow{2}{*}{YOLOv5x}\cite{ultralytics_yolov5} & RGB & 0.82 & 0.72 & 0.68 & 0.59 & 0.48 & 0.66 & 0.39 & 0.62 & 0.62\\
        & RGB + IR & 0.84 & 0.73 & 0.70 & 0.61 & 0.50 & 0.67 & 0.39 & 0.57 & 0.62\\
        \midrule
        
        \multirow{2}{*}{YOLOrs}\cite{sharma2020yolors} & RGB & \textbf{0.85} & 0.73 & 0.70 & 0.51 & 0.43 & \textbf{0.77} & 0.19 & 0.39 & 0.57\\
        & RGB +IR & 0.84 & 0.78 & 0.69 & 0.53 & 0.47 & 0.68 & 0.21 & 0.58 & 0.60\\
        
        
        \midrule
        \multirow{2}{*}{Ours}
        & RGB & 0.81 & 0.74 & 0.64 & 0.63 & \textbf{0.53} & 0.63 & 0.53 & 0.61 & 0.64\\
        & RGB + IR & 0.80 & \textbf{0.79} & 0.70 & \textbf{0.73} & 0.45 & 0.72 & \textbf{0.53} & \textbf{0.68} & \textbf{0.68} \\
        \bottomrule
    \end{tabular}
    \end{adjustbox}

    \label{tab:globalcomparaison}
\end{table*}

\subsubsection{Ablation study}
We verify the effectiveness of our proposed method by designing a series of ablation experiments conducted on the first fold of the validation set of VEDAI.

\noindent\textbf{a) Effectiveness of cross-attention MIM} 
Table \ref{table:EffectivenessCA-RGB} shows the effectiveness of the proposed interactive MIM using cross-attention for pre-training on RGB images. To provide a baseline for comparison, we first trained a model from scratch on VEDAI for object detection without any pre-training, achieving a score of 0.50 for mAP@.5. Next, we applied conventional MIM such as MAE, to pre-train the encoder on VEDAI and fine-tune the pre-trained encoder for the downstream task. This resulted in a subpar score of 0.42, confirming that SSL requires a large dataset to be effective \cite{chen2020simple}. The score improved to 0.52 when pre-training the encoder on the larger DIOR dataset (20x larger than VEDAI). The performance improves from 0.52 to 0.62, when we use the proposed interactive MIM introducing a cross-attention module into the standard MIM framework. This improvement demonstrates that the interaction between unmasked tokens and masked tokens enables the encoder to learn fine-grained details in the surrounding context that are crucial for detecting small or densely packed objects in remote sensing images. Additionally, we also use unmasked tokens as the query matrix Q to generate its features through the cross-attention module. As expected, the performance was lower than the previous approach since partial information is missing when using the theirs features instead of original unmasked tokens.

\begin{table}[htbp]
    \caption{Ablation study on the effectiveness of interactive MIM using cross-attention as pre-training for the object detection task.}
    
    \resizebox{\columnwidth}{!}{\begin{tabular}{cccccccccccc}
    
    \toprule
    \multirow{2}{*}{\textbf{Modality}}&\multirow{2}{*}{\textbf{mAP@.5}}  &  \multicolumn{2}{c}{\textbf{pre-training datasets}} & \multicolumn{2}{c}{\textbf{Cross Attention}}\\
            \cmidrule(lr){3-4} \cmidrule(lr){5-6}
            & & {VEDAI} & {DIOR} & {Q masked} & {Q unmasked} \\
    \midrule
    \multirow{5}{*}{RGB}
                        &0.50        &(Training from scratch, w/o pre-training) & & & \\
                        &0.42       & \checkmark & & & \\
                        &0.52       & & \checkmark & \\
                        &\textbf{0.62}       & & \checkmark & \checkmark & \\
                        &0.60       & & \checkmark & & \checkmark \\
    \bottomrule
    \end{tabular}}
    \label{table:EffectivenessCA-RGB}
\end{table}

\noindent\textbf{b) Effectiveness of multimodal interactive MIM} 
Table \ref{table:EffectivenessCA-Multimodal} also shows the effectiveness of the proposed interactive MIM when using multimodal images. Compared to the baseline only uses RGB images, the new baseline using multiple modalities significantly improves from 0.50 to 0.63 in mAP@.5, demonstrating the effectiveness of multimodal learning for the object detection in remote sensing. Moreover, the performance further improves to 0.64 when using proposed interactive MIM. However, the AVIID dataset contains only 1246 images, which limits the power of the proposed method. We augmented the data by resizing the images from 480x480 pixels to 512x512 pixels and then splitting them into 4256 images. Then the performance improves to 0.68, demonstrating that our proposed method is also effective on multimodal images. 

\begin{table}[htbp]
    \caption{Ablation study of using interactive MIM as pre-training for the multimodal object detection task. }
    
    \resizebox{\columnwidth}{!}{\begin{tabular}{cccccccccccc}
    
    \toprule
    \multirow{2}{*}{\textbf{Modality}}&\multirow{2}{*}{\textbf{mAP@.5}}  &  \multicolumn{2}{c}{\textbf{AVIID}} & \multicolumn{1}{c}{\textbf{Cross Attention}}\\
            \cmidrule(lr){3-4} \cmidrule(lr){5-6}
            & & {original} & {data expansion} & {Q masked}  \\
    \midrule
    \multirow{3}{*}{RGB + IR}
                        &0.63 &(Training from scratch, w/o pre-training) & & & \\
                        &0.64 & \checkmark &  & \checkmark & \\
                        &\textbf{0.68} & & \checkmark & \checkmark & \\
    \bottomrule
    \end{tabular}}
    \label{table:EffectivenessCA-Multimodal}
\end{table}

\noindent\textbf{c) Impact of mask size} 
The varying size of objects, especially small objects, is a major challenge in object detection within remote sensing images. The impact of applying different mask sizes in pre-training, based on our proposed method, on the downstream object detection task is shown in Table \ref{table:maskz}. The results indicate that the optimal performance was achieved with medium-sized masks of 32x32 pixels, offering a balance between capturing global context with large masks and localized fine-grained feature learning with small masks. The 32x32 pixel is the default mask size used in our experiments.

\begin{table}[h!]
    \caption{Impact of applying masks with different sizes in pre-training    on downstream object detection task.}
    \centering
    \begin{tabular}{c|c|c|c}
    \toprule
    \textbf{Mask size} & 16 & 32 & 64  \\
    \midrule
    \textbf{mAP@.5}  & 0.63 & \textbf{0.68} & 0.61\\
    \bottomrule
    \end{tabular}
    \label{table:maskz}
    \end{table}
\begin{figure}
    \centering
    \includegraphics[width=\linewidth]{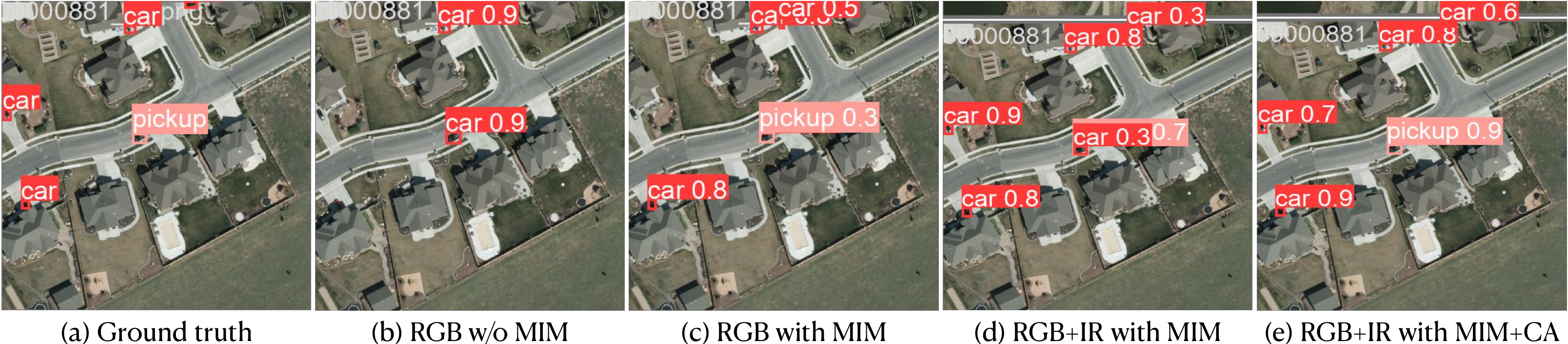}
    \caption{Visual illustration of the effectiveness of the proposed interactive MIM for multimodal object detection in remote sensing images.}
    \label{fig:visual_result}
\end{figure}
\subsubsection{Overall evaluation}
Table \ref{tab:globalcomparaison} compares our proposed method with other methods on the object detection task on the benchmark VEDAI dataset. Our model demonstrates superiority in both single modality using RGB images and multimodality using RGB+IR images. Notably, our model also proves effective in detecting pickups both in single moality, which are easily confused with trucks, and in detecting relatively small objects, such as boats.


\subsubsection{Visualization}
Figure \ref{fig:visual_result} visually demonstrates the effectiveness of the multimodal cross-attention MIM for the object detection task. In (b), compared to the ground truth shown in (a), several objects, such as the pickup and cars, are missing when using only RGB images without pre-training. With pre-training based on MIM, the pickup and more cars are detected, as shown in (c). When using multimodal images (RGB+IR) with conventional MIM, all the cars are detected, but the car and pickup are still confused. Finally, when using multimodal RGB+IR images with cross-attention MIM, all objects are successfully detected with high accuracy.

\section{Conclusion}
The results demonstrate that integrating self-supervised learning via MIM and multimodal data fusion significantly improves the performance of object detection in remote sensing images. Our experiments show that incorporating IR data alongside RGB enhances the model's capability, especially for small and occluded objects. The proposed interactive MIM, which introduces a cross-attention module to establishe the interaction between unmasked and masked tokens, overcomes the shortcomings of conventional MIM. This approach encourages the encoder to learn a more holistic understanding of the image content, benefiting downstream object detection tasks in remote sensing images. Although our work is initially proposed for remote sensing, it can also be extended to other domains, applying to more scenarios with limited resources.



\bibliographystyle{ieeetr}
\bibliography{refs}

\end{document}